\documentclass[letterpaper, 10 pt, conference]{ieeeconf}  

\IEEEoverridecommandlockouts                              
\makeatletter
\let\NAT@parse\undefined
\makeatother

\usepackage{amsmath} 
\usepackage{amssymb}  

\usepackage{array}
\usepackage{float}
\usepackage{multirow}
\usepackage{graphicx}
\usepackage{subfig}
\usepackage{algorithmic}
\usepackage[square, comma, sort&compress, numbers]{natbib}

\let\vec\boldvec

\providecommand{\tabularnewline}{\\}
\floatstyle{ruled}
\newfloat{algorithm}{tbp}{loa}
\providecommand{\algorithmname}{Algorithm}
\floatname{algorithm}{\protect\algorithmname}

\title{\LARGE \bf
Learning Prioritized Control of Motor Primitives*
}

\author{Jens Kober and Jan Peters
\thanks{*The project receives funding from the European Community's Seventh
Framework Programme under grant agreement no. ICT-248273 GeRT. The
project receives funding from the European Community's Seventh Framework
Programme under grant agreement no. ICT-270327 CompLACS.}
\thanks{Both authors are with the Max Planck Institute for Intelligent Systems,
Department of Empirical Inference, Spemannstr. 38, 72076 T\"ubingen,
Germany and the Technische Universit\"at Darmstadt, Intelligent Autonomous
Systems Group, Hochschulstr. 10, 64289 Darmstadt, Germany {\tt\small \{firstname.lastname\}@tuebingen.mpg.de}}%
}

\begin{document}

\maketitle
\thispagestyle{empty}
\pagestyle{empty}

\begin{abstract}

Many tasks in robotics can be decomposed into sub-tasks that are performed simultaneously. In many cases, these sub-tasks cannot all be achieved jointly and a prioritization of such sub-tasks is required to resolve this issue. In this paper, we discuss a novel learning approach that allows to learn a prioritized control law built on a set of sub-tasks represented by motor primitives. The primitives are executed simultaneously but have different priorities. Primitives of higher priority can override the commands of the conflicting lower priority ones. The dominance structure of these primitives has a significant impact on the performance of the prioritized control law. We evaluate the proposed approach with a ball bouncing task on a Barrett WAM.

\end{abstract}

\section{INTRODUCTION}

When learning a new skill, it is often easier to practice the required
sub-tasks separately and later on combine them to perform the task
-- instead of attempting to learn the complete skill as a whole. For
example, in sports sub-tasks can often be trained separately. Individual
skills required in the sport are trained in isolation to improve the
overall performance, e.g., in volleyball a serve can be trained without
playing the whole game. 

Sub-tasks often have to be performed simultaneously and it is not
always possible to completely fulfill all at once. Hence, the sub-tasks
need to be prioritized. An intuitive example for this kind of prioritizing
sub-tasks happens during a volleyball game: a player considers hitting
the ball (and hence avoiding it touching the ground and his team loosing
a point) more important than locating a team mate and playing the
ball precisely to him. The player will attempt to fulfill both sub-tasks.
If this is not possible it is often better to ``save'' the ball
with a high hit and hope that another player recovers it rather than
immediately loosing a point.

\begin{figure}
\centering{}\includegraphics[width=0.8\columnwidth]{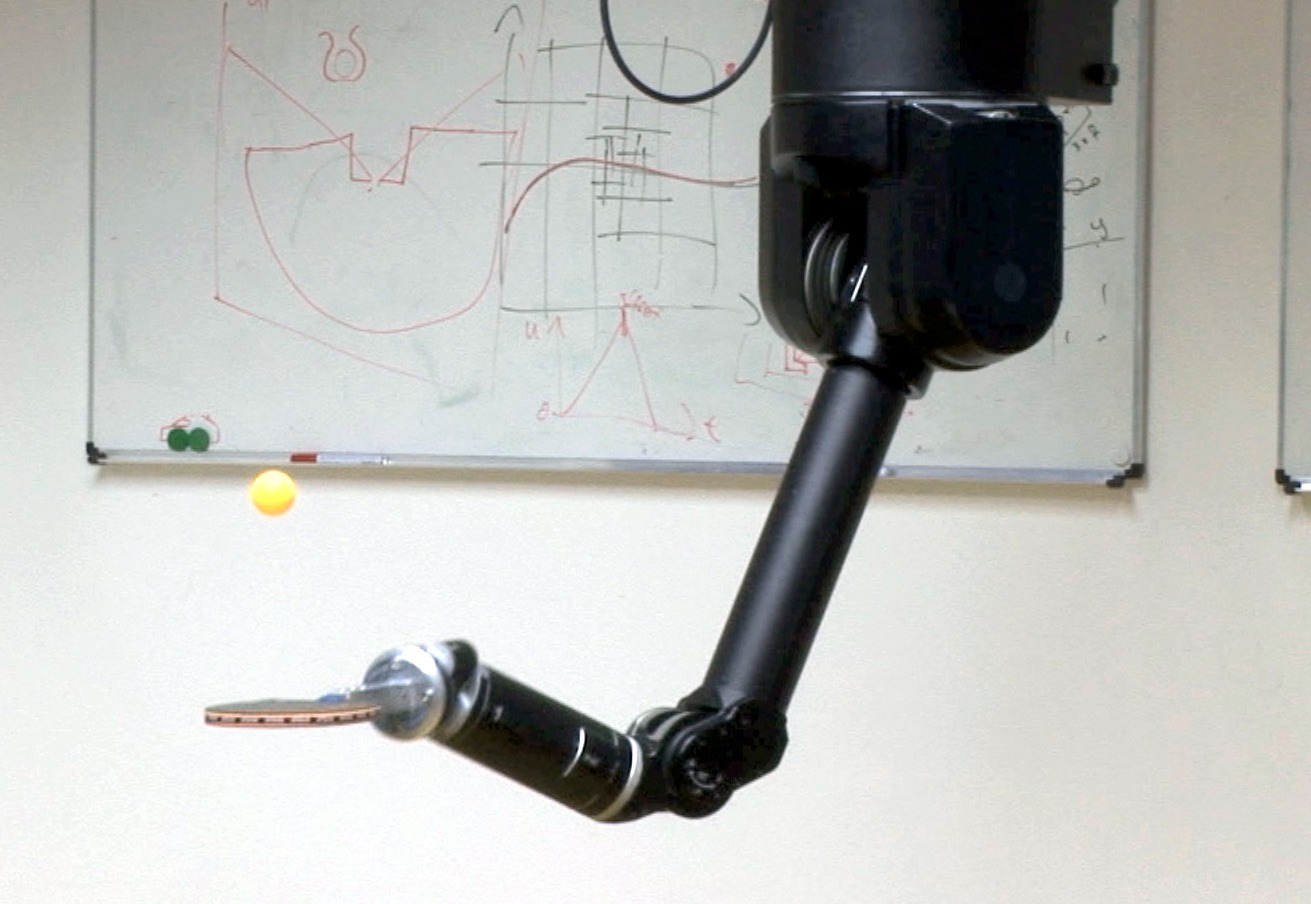}\caption{\label{fig:wam}This figure illustrates the ball-bouncing task on
a Barrett WAM. The goal is to keep the ball bouncing on the racket.}
\end{figure}

In this paper, we learn different sub-tasks that are represented by
motor primitives that combined can perform a more complicated task.
For doing so, we will stack controls corresponding to different primitives
that represent movements in task space. These primitives are assigned
different priorities and the motor commands corresponding to primitives
with higher priorities can override the motor commands of lower priority
ones. The proposed approach is outlined in Sect.~\ref{sub:Proposed-Approach}
and further developed in Sect.~\ref{sec:Learning-the-Hierarchical}.
We evaluate our approach with a ball-bouncing task (see Fig.~\ref{fig:wam}
and Sect.~\ref{sec:Evaluation:-Ball-Bouncing}).

As the sub-tasks describe the movements in task space, we have to
learn a control that is mapping to the robot joint space. Unfortunately,
this mapping is not a well-defined function for many robots. For example,
if the considered task space has fewer degrees of freedom than the
robot, multiple solutions are possible. This redundancy can be resolved
by introducing a null-space control, i.e., a behavior that operates
on the redundant degrees of freedom. Such a null-space control can
for example pull the robot towards a rest posture \cite{Peters2008},
prevent getting close to joint limits \cite{Chaumette2001}, avoid
obstacles \cite{Khatib:1986nj} or singularities \cite{Yoshikawa1985}.
Computing the task space control often corresponds to an optimization
problem, that can for example be solved by a gradient based approach.
A well known approach is the pseudo-inverse solution \cite{Khatib:1986nj,Peters2008}.
An alternative is to learn an operational space control law that implicitly
includes the null-space behavior \cite{Peters2008a}. Once learned,
it corresponds to a unique mapping from desired actions in operational
space to required actions in joint space.

The problem studied in this paper is related to hierarchical control
problems as discussed in \cite{Findeisen1980}. Using prioritized
primitives in classical control has been explored in \cite{Sentis05a}
by using analytical projections into the null-space. In this paper,
we propose a learning approach that does not require complete knowledge
of the system, the constraints, and the task. In the reinforcement
learning community, the compositions of options (i.e., concurrent
options), which is related to the concurrent execution of primitives,
has been studied \cite{Precup1998}. Learning null-space control has
been explored in \cite{Towell2010}. In contrast, we do not attempt
to recover the implicit null-space policy but build a hierarchical
operational space control law from user demonstrated primitives.

\subsection{Proposed Approach\label{sub:Proposed-Approach}}

Based on the observation that many tasks can be described as a superposition
of sub-tasks, we want to have a set of controls that can be executed
simultaneously. As a representation for the sub-tasks, we chose the
dynamical systems motor primitives, which are discussed in more detail
in Section~\ref{sub:BackgroundDMPs}. Such primitives are well suited
as representation for the sub-tasks as they ensure the stability of
the movement generation. They are invariant under transformations
of the initial position and velocity, the final position and velocity,
the duration as well as the movement amplitude.

In this paper, these primitives are described in different task spaces,
e.g., in the form 
\[
\ddot{\vec{x}}_{i}=\pi_{i}(\vec{x}_{i},\dot{\vec{x}}_{i},z)
\]
where $z$ denotes a shared canonical system while $\vec{x}_{i}$ are
positions in task-space $i$. For example, if we have a primitive
``move end-effector up and down'' its task space would correspond
to the Cartesian position indicating the height (as well as the corresponding
velocities and accelerations) but not include the sideways movement
or the orientation of the end-effector. The dynamical systems motor
primitives are well suited to represent different kinds of vertical
movements starting and ending at various states and of different duration.

These primitives are prioritized such that 
\[
i\succeq i-1,
\]
which reads a ``task $i$ dominates task $i-1$''. If both sub-tasks
can be fulfilled at the same time, our system will do so -- but if
this should not be possible, sub-task $i$ will be fulfilled at the
expense of sub-task $i-1$. We attempt to reproduce a complex task
that consists of several sub-tasks, represented by motor primitives,
\[
\{\pi_{1},\pi_{2,},\ldots,\pi_{N}\}
\]
 that are concurrently executed at the same time following the prioritization
scheme
\[
N\succeq N-1\succeq\cdots\succeq2\succeq1.
\]
This approach requires a prioritized control law that composes the
motor command out of the primitives $\pi_{i}$, i.e., 
\[
\vec{u}=\vec{f}(\pi_{1},\pi_{2,},\ldots,\pi_{N},\vec{q},\dot{\vec{q}})
\]
where $\vec{q},\vec{\dot{q}}$ are the joint position and joint velocity,
$\vec{u}$ are the generated motor commands (torques or accelerations).

We try to acquire the prioritized control law in three steps, which
we will illustrate with the ball-bouncing task: 
\begin{enumerate}
\item We observe $\ddot{\vec{x}}_{i}(t),\dot{\vec{x}}_{i}(t),\vec{x}_{i}(t)$
individually for each of the primitives that will be used for the
task. For the ball-bouncing example, we may have the following sub-tasks:
``move under the ball'', ``hit the ball'', and ``change racket
orientation''. The training data is collected by executing only one
primitive at a time without considering the global strategy, e.g.,
for the ``change racket orientation'' primitive by keeping the position
of the racket fixed and only changing its orientation without a ball
being present. This training data is used to acquire the task by imitation
learning under the assumption that these tasks did not need to overrule
each other in the demonstration (Sect.~\ref{sec:Learning-the-Hierarchical}).
\item We enumerate all possible dominance structures and learn a prioritized
control law for each dominance list that fuses the motor primitives.
For the three ball-bouncing primitives there are six possible orders,
as listed in Table~\ref{tab:dominance}. 
\item We choose the most successful of these approaches. The activation
and adaptation of the different primitives is handled by a strategy
layer (Sect.~\ref{sub:Strategy}). In the ball-bouncing task, we
evaluate how long each of the prioritized control laws keeps the ball
in the air and pick the best performing one (Sect.~\ref{sub:Learning-Results}).
\end{enumerate}
Clearly, enumerating all possible dominance structures only works
for small systems (as the number of possibilities grows with $n!$,
i.e., exponentially fast).

\section{BACKGROUND: MOTOR PRIMITIVES\label{sub:BackgroundDMPs}}

While the original formulation in \cite{ADM:Ijs2002} for discrete
dynamical systems motor primitives used a second-order system to represent
the phase $z$ of the movement, this formulation has proven to be
unnecessarily complicated in practice. Since then, it has been simplified
and, in \cite{Schaal2007}, it was shown that a single first order
system suffices 
\begin{equation}
\dot{z}=-\tau\alpha_{z}z.\label{eq:canonical-impl_d}
\end{equation}
 This canonical system has the time constant $\tau=1/T$ where $T$
is the duration of the motor primitive, a parameter $\alpha_{z}$
which is chosen such that $z\approx0$ at $T$ to ensure that the
influence of the transformation function, shown~in Eq. (\ref{eq:trafo:func_d-1}),
vanishes. Subsequently, the internal state $\vec{y}$ of a second
system is chosen such that positions $\vec{x}$ of all degrees of
freedom are given by $\vec{x}=\vec{y}_{1}$, the velocities $\dot{\vec{x}}$
by $\dot{\vec{x}}=\tau\vec{y}_{2}=\dot{\vec{y}}_{1}$ and the accelerations
$\ddot{\vec{x}}$ by $\ddot{\vec{x}}=\tau\dot{\vec{y}}_{2}$. Under
these assumptions, the learned dynamics of Ijspeert motor primitives
can be expressed in the following form

\begin{align}
\dot{\vec{y}}_{2} & =\tau\alpha_{y}\left(\beta_{y}\left(\vec{g}-\vec{y}_{1}\right)-\vec{y}_{2}\right)+\tau\vec{A}\vec{f}\left(z\right),\label{eq:implementation1_d}\\
\dot{\vec{y}}_{1} & =\tau\vec{y}_{2}.\nonumber 
\end{align}
 This set of differential equations has the same time constant $\tau$
as the canonical system, parameters $\alpha_{y}$, $\beta_{y}$ set
such that the system is critically damped, a goal parameter $\vec{g}$,
a transformation function $\vec{f}$ and an amplitude matrix $\vec{A}=\operatorname{diag}(a_{1},a_{2},\ldots,a_{n})$,
with the amplitude modifier $\vec{a}=[a_{1},a_{2},\ldots,a_{n}]$.
In \cite{Schaal2007}, they use $\vec{a}=\vec{g}-\vec{y}_{1}^{0}$
with the initial position $\vec{y}_{1}^{0}$, which ensures linear
scaling. Alternative choices are possibly better suited for specific
tasks, see e.g., \cite{DaeHyungPark_humanoids_2008}. The transformation
function $\vec{f}\left(z\right)$ alters the output of the first system,
in Eq.~(\ref{eq:canonical-impl_d}), so that the second system, in
Eq.~(\ref{eq:implementation1_d}), can represent complex nonlinear
patterns and it is given by
\begin{equation}
\vec{f}\left(z\right)={\textstyle \sum}_{i=1}^{N}\psi_{i}\left(z\right)\vec{w}_{i}z.\label{eq:trafo:func_d-1}
\end{equation}
 Here, $\vec{w}_{i}$ contains the $i^{\text{th}}$ adjustable parameter
of all degrees of freedom, $N$ is the number of parameters per degree
of freedom, and $\psi_{i}(z)$ are the corresponding weighting functions
\cite{Schaal2007}. Normalized Gaussian kernels are used as weighting
functions given by
\[
\psi_{i}\left(z\right)=\frac{\exp\left(-h_{i}\left(z-c_{i}\right)^{2}\right)}{\sum_{j=1}^{N}\exp\left(-h_{j}\left(z-c_{j}\right)^{2}\right)}.
\]
 These weighting functions localize the interaction in phase space
using the centers $c_{i}$ and widths $h_{i}$. Note that the degrees
of freedom (DoF) are usually all modeled as independent in Eq.~(\ref{eq:implementation1_d}).
All DoFs are synchronous as the dynamical systems for all DoFs start
at the same time, have the same duration, and the shape of the movement
is generated using the transformation $\vec{f}\left(z\right)$ in
Eq.~(\ref{eq:trafo:func_d-1}). This transformation function is learned
as a function of the shared canonical system in Eq.~(\ref{eq:canonical-impl_d}).

The original formulation assumes that the goal velocity is zero. Clearly
this behavior is undesirable for hitting the balls in the ball-bouncing
task. In \cite{Kober2010}, we proposed a modification that allows
to specify arbitrary goal velocities: 
\begin{align*}
\dot{\vec{y}}_{2} & =\left(1-z\right)\tau\alpha_{g}\left(\beta_{g}\left(\vec{g}_{m}-\vec{y}_{1}\right)+\frac{\left(\dot{\vec{g}}-\dot{\vec{y}}_{1}\right)}{\tau}\right)+\tau\vec{A}\vec{f}\\
\dot{\vec{y}}_{1} & =\tau\vec{y}_{2},\\
\vec{g}_{m} & =\vec{g}_{m}^{0}-\dot{\vec{g}}\frac{\ln\left(z\right)}{\tau\alpha_{h}},
\end{align*}
where $\dot{\vec{g}}$ is the desired final velocity, $\vec{g}_{m}$
is the moving goal and the initial position of the moving goal $\vec{g}_{m}^{0}=\vec{g}-\tau\dot{\vec{g}}$
ensures that $\vec{g}_{m}\left(T\right)=\vec{g}$. The term $-\ln\left(z\right)/\left(\tau\alpha_{h}\right)$
is proportional to the time if the canonical system in Eq.~(\ref{eq:canonical-impl_d})
runs unaltered; however, adaptation of $z$ allows the straightforward
adaptation of the hitting time.

As suggested in \cite{ADM:Ijs2002}, locally-weighted linear regression
can be used for imitation learning. The duration of discrete movements
is extracted using motion detection and the time-constants are set
accordingly. Additional feedback terms can be added as shown in \cite{DaeHyungPark_humanoids_2008,ADM:Ijs2002,Schaal2007}.

\section{LEARNING THE PRIORITIZED CONTROL LAW\label{sec:Learning-the-Hierarchical}}

By learning the prioritized control, we want to obtain a control law

\[
\vec{u}=\ddot{\vec{q}}=\vec{f}(\pi_{1},\pi_{2,},\ldots,\pi_{N},\vec{q},\dot{\vec{q}}),
\]
i.e., we want to obtain the required control $\vec{u}$ that executes
the primitives $\pi_{1},\pi_{2,},\ldots,\pi_{N}$. Here, the controls
correspond to the joint accelerations $\ddot{\vec{q}}$. The required
joint accelerations not only depend on the primitives but also on
the current state of the robot, i.e., the joint positions $\vec{q}$
and joint velocities $\dot{\vec{q}}$. Any control law can be represented
locally as a linear control law. In our setting, these linear control
laws can be represented as 
\[
\vec{u}=\left[\begin{array}{c}
\ddot{\vec{x}}_{i}\\
\dot{\vec{q}}\\
\vec{q}
\end{array}\right]^{\mathrm{T}}\vec{\theta}=\vec{\phi}^{\mathrm{T}}\vec{\theta},
\]
where $\vec{\theta}$ are the parameters we want to learn and $\vec{\phi}=\left[\begin{array}{ccc}
\ddot{\vec{x}}_{i} & \vec{\dot{q}} & \vec{q}\end{array}\right]$ acts as features. Often the actions of the primitive $\ddot{\vec{x}}_{i}$
can be achieved in multiple different ways due to the redundancies
in the robot degrees of freedom. To ensure consistency, a null-space
control is introduced. The null-space control can, for example, be
defined to pull the robot towards a rest posture $\vec{q}_{0}$, resulting
in the null-space control 
\[
\vec{u}_{0}=-\vec{K}_{D}\dot{\vec{q}}-\vec{K}_{P}\left(\vec{q}-\vec{q}_{0}\right),
\]
where $\vec{K}_{D}$ and $\vec{K}_{P}$ are gains for the velocities
and positions respectively. 

To learn the prioritized control law, we try to generalize the learning of the
operational space control approach from \cite{Peters2008a} to a hierarchical
control approach \cite{Sentis05a,Peters2008}.

\subsection{Single Primitive Control Law\label{sub:Single-Control-Law}}

A straightforward approach to learn the motor commands $\vec{u}$,
represented by the linear model $\vec{u}=\vec{\phi}^{\mathrm{T}}\vec{\theta}$,
is using linear regression. This approach minimizes the squared error
\[
\mathrm{E}^{2}=\sum_{t=1}^{T}\left(\vec{u}_{t}^{\textrm{ref}}-\vec{\phi}_{t}^{\mathrm{T}}\vec{\theta}\right)^{2}
\]
between the demonstrated control of the primitive $u_{t}^{\textrm{ref}}$
and the recovered linear policy $\vec{u}_{t}=\vec{\phi}_{t}^{\mathrm{T}}\vec{\theta}$,
where $T$ is the number of samples. The parameters minimizing this
error are 
\begin{equation}
\vec{\theta}=\left(\vec{\Phi}^{T}\vec{\Phi}+\lambda\vec{I}\right)^{-1}\vec{\Phi}^{\mathrm{T}}\vec{U},\label{eq:regression}
\end{equation}
with $\vec{\Phi}$ and $\vec{U}$ containing the values of the demonstrated
$\vec{\phi}$ and $\vec{u}$ for all time-steps $t$ respectively,
and a ridge factor $\lambda$. If the task space and the joint-space
coincide, the controls $\vec{u}=\ddot{\vec{q}}$ are identical to
the action of the primitive $\ddot{\vec{x}}_{i}$. We also know that
\emph{locally} any control law that can be learned from data is a
viable control law \cite{Peters2008a}. The error with respect to
the training data is minimized, \emph{however}, if the training data
is not consistent, the plain linear regression will average the motor
commands, which is unlikely to fulfill the actions of the primitive.

In order to enforce consistency, the learning approach has to resolve
the redundancy and incorporate the null-space control. We can achieve
this by using the program 
\begin{eqnarray}
\min_{\vec{u}}J & = & \left(\vec{u}-\vec{u}_{0}\right)^{\mathrm{T}}\vec{N}\left(\vec{u}-\vec{u}_{0}\right)\label{eq:program}\\
s.t.\:\ddot{\vec{x}} & = & \pi\left(\vec{x},\dot{\vec{x}},z\right)\nonumber 
\end{eqnarray}
as discussed in \cite{Peters2008}. Here the cost $J$ is defined
as the weighted squared difference of the control $\vec{u}$ and the
null-space control $\vec{u}_{0}$, where the metric $\vec{N}$ is
a positive semi-definite matrix. The idea is to find controls $\vec{u}$
that are as close as possible to the null-space control $\vec{u}_{0}$
while still fulfilling the constraints of the primitive $\pi$. This
program can also be solved as discussed in \cite{Peters2008a}. Briefly
speaking, the regression in Eq.~(\ref{eq:regression}) can be made
consistent by weighting down the error by weights $w_{t}$ and hence
obtaining 
\begin{equation}
\vec{\theta}=\left(\vec{\Phi}^{\mathrm{T}}\vec{W}\vec{\Phi}+\lambda\vec{I}\right)^{-1}\vec{\Phi}^{\mathrm{T}}\vec{W}\vec{U}\label{eq:RWR}
\end{equation}
with $\vec{W}=\operatorname{diag}(w_{1},\dots,w_{Tn})$ for $T$ samples.
This approach works well for linear models and can be gotten to work
with multiple locally linear control laws. Nevertheless, it maximizes
a reward instead of minimizing a cost. The cost $J$ can be transformed
into weights $w_{t}$ by passing it through an exponential function
\[
w_{t}=\exp\left(-\alpha\tilde{\vec{u}}_{t}^{\mathrm{T}}\vec{N}\tilde{\vec{u}}_{t}\right),
\]
where $\tilde{\vec{u}}_{t}=\left(\vec{u}_{t}-\vec{u}_{0}\right)$.
The scaling factor $\alpha$ acts as a monotonic transformation that
does not affect the optimal solution but can increase the efficiency
of the learning algorithm.

\begin{algorithm}
\caption{\label{alg:Learning-the-Prioritized}Learning the Prioritized Control
Law}
\begin{algorithmic} 

\STATE define null-space control $\vec{u}_{0}$, metric $\vec{N}$,
scaling factor $\alpha$ 

\medskip{}

\STATE collect controls $\vec{u}_{i,t}$ and features $\vec{\phi}_{i,t}^{\mathrm{}}$
for all primitives $i\in\left\{ 1,\ldots,N\right\} $ and all time-steps
$t\in\left\{ 1,\ldots,T\right\} $ separately \medskip{}

\FOR{primitives $i=1\ldots N$ {\footnotesize (N: highest priority)}}\medskip{}

\FOR{time-steps $t=1\ldots T$ }\medskip{}

\STATE calculate offset controls\\
$\hat{\vec{u}}_{i,t}=\vec{u}_{i,t}-\sum_{j=1}^{i-1}\vec{\phi}_{i,t}^{\mathrm{T}}\vec{\theta}_{j}-\vec{u}_{0,t}$\medskip{}

\STATE calculate weights $\hat{w}_{i,t}=\exp\left(-\alpha\hat{\vec{u}}_{i,t}^{\mathrm{T}}\vec{N}\hat{\vec{u}}_{i,t}\right)$
\medskip{}

\ENDFOR\medskip{}

\STATE  build control matrix $\hat{\vec{U}}_{i}$ containing $\hat{\vec{u}}_{i,1}\ldots\hat{\vec{u}}_{i,T}$\medskip{}

\STATE  build feature matrix $\vec{\Phi}_{i}$ containing $\vec{\phi}_{i,1}\ldots\vec{\phi}_{i,T}$\medskip{}

\STATE  build weight matrix $\hat{\vec{W}}_{i}=\operatorname{diag}(\hat{w}_{i,1},\dots,\hat{w}_{i,T})$ 

\medskip{}
\STATE  calculate parameters\\
$\vec{\theta}_{i}=\left(\vec{\Phi}_{i}^{\mathrm{T}}\hat{\vec{W}}_{i}\vec{\Phi}_{i}+\lambda\vec{I}\right)^{-1}\vec{\Phi}_{i}^{\mathrm{T}}\hat{\vec{W}}_{i}\hat{\vec{U}}_{i}$\medskip{}

\ENDFOR

\end{algorithmic} \textbf{end for} 
\end{algorithm}
Using the Woodbury formula \cite{Welling2010} Eq.~(\ref{eq:RWR})
can be transformed into 
\begin{equation}
\vec{u}=\vec{\phi}(x)^{\mathrm{T}}\vec{\Phi}^{\mathrm{T}}\left(\vec{\Phi}\vec{\Phi}^{\mathrm{T}}+\vec{W}_{U}\right)^{-1}\vec{U}\label{eq:kernel}
\end{equation}
with $\vec{W}_{U}=\textrm{diag}\left(\tilde{\vec{u}}_{1}^{T}\vec{N}\tilde{\vec{u}}_{1},\dots,\tilde{\vec{u}}_{n}^{T}\vec{N}\tilde{\vec{u}}_{n}\right)$.
By introducing the kernels $\vec{k}(\vec{s})=\vec{\phi}(\vec{s})^{\mathrm{T}}\vec{\Phi}^{\mathrm{T}}$
and $\vec{K}=\vec{\Phi}\vec{\Phi}^{\mathrm{T}}$ we obtain 
\[
\vec{u}=\vec{k}(\vec{s})^{\mathrm{T}}\left(\vec{K}+\vec{W}_{U}\right)^{-1}\vec{U},
\]
which is related to the kernel regression \cite{Bishop:2006rm}. This
kernelized form of Eq.~(\ref{eq:kernel}) overcomes the limitations
of the linear model at a cost of higher computational complexity.

\subsection{Prioritized Primitives Control Law\label{sub:Prioritized-Primitives-Control}}

In the previous section, we have described how the control law for
a single primitive can be learned. To generalize this approach to
multiple primitives with different priorities, we want a control law
that always fulfills the primitive with the highest priority and follows
the remaining primitives as much as possible according to their place
in the hierarchy. Our idea is to represent the higher priority control
laws as correction term with respect to the lower priority primitives.
The control of the primitive with the lowest priority is learned first.
This control is subsequently considered to be a baseline and the primitives
of higher priority only learn the difference to this baseline control.
The change between the motor commands resulting from primitives of
lower priority is minimized. The approach is reminiscent of online
passive-aggressive algorithms \cite{Crammer2006}. Hence, control
laws of higher priority primitives only learn the offset between their
desired behavior and the behavior of the lower priority primitives.
This structure allows them to override the actions of the primitives
of lesser priority and, therefore, add more detailed control in the
regions of the state space they are concerned with. The combined control
of all primitives is
\[
\vec{u}=\vec{u}_{0}+\sum_{n=1}^{N}\Delta\vec{u}_{n},
\]
 where $\vec{u}_{0}$ is the null-space control and $\Delta\vec{u}_{n}$
are the offset controls of the $N$ primitives.

Such control laws can be expressed by changing the program in Eq.~(\ref{eq:program})
to
\begin{eqnarray*}
\min_{\vec{u}_{i}}J & \!\!= & \!\!\!\!\left(\vec{u}_{i}-\sum_{j=1}^{i-1}\Delta\vec{u}_{j}-\vec{u}_{0}\right)^{\mathrm{T}}\!\!\!\vec{N}\!\left(\vec{u}_{i}-\sum_{j=1}^{i-1}\Delta\vec{u}_{j}-\vec{u}_{0}\right)\\
s.t.\:\ddot{\vec{x}}_{i} & \!\!= & \!\pi_{i}(\vec{x}_{i},\dot{\vec{x}}_{i},z),
\end{eqnarray*}

where the primitives need to be learned in the increasing order of
their priority, the primitive with the lowest priority is learned
first, the primitive with the highest priority is learned last. The
regression in Eq.~(\ref{eq:RWR}) changes to 
\[
\vec{\theta}_{i}=\left(\vec{\Phi}_{i}^{\mathrm{T}}\hat{\vec{W}}_{i}\vec{\Phi}_{i}+\lambda\vec{I}\right)^{-1}\vec{\Phi}_{i}^{\mathrm{T}}\hat{\vec{W}}_{i}\hat{\vec{U}}_{i},
\]
where $\hat{\vec{U}}_{i}$ contains the offset controls $\hat{\vec{u}}_{i,t}=\vec{u}_{i,t}-\sum_{j=1}^{i-1}\Delta\vec{u}_{j,t}-\vec{u}_{0,t}$
for all time-steps $t$, where $\Delta\vec{u}_{j,t}=\vec{\phi}_{i,t}^{\mathrm{T}}\vec{\theta}_{j}$.
The weighting matrix $\hat{\vec{W}}_{i}$ now has the weights $\hat{w}_{t}=\exp\left(-\alpha\hat{\vec{u}}_{i,t}^{\mathrm{T}}\vec{N}\hat{\vec{u}}_{i,t}\right)$
on its diagonal and matrix $\hat{\vec{U}}_{i}$ contains offset controls
$\hat{\vec{u}}_{i,t}$. The kernelized form of the prioritized control
law can be obtained analogously. The complete approach is summarized
in Algorithm~\ref{alg:Learning-the-Prioritized}.

\section{EVALUATION: BALL-BOUNCING\label{sec:Evaluation:-Ball-Bouncing}}

In order to evaluate the proposed prioritized control approach, we
chose a ball bouncing task. We describe the task in Section~\ref{sub:Task-Description},
explain a possible higher level strategy in Section~\ref{sub:Strategy},
and discuss how the proposed framework can be applied in Section~\ref{sub:Learning-Results}.

\subsection{Task Description\label{sub:Task-Description}}

\begin{figure*}
\subfloat[\label{fig:schematic_drawing}Exaggerated schematic drawing. The green
arrows indicate velocities.]{\includegraphics[width=1\textwidth]{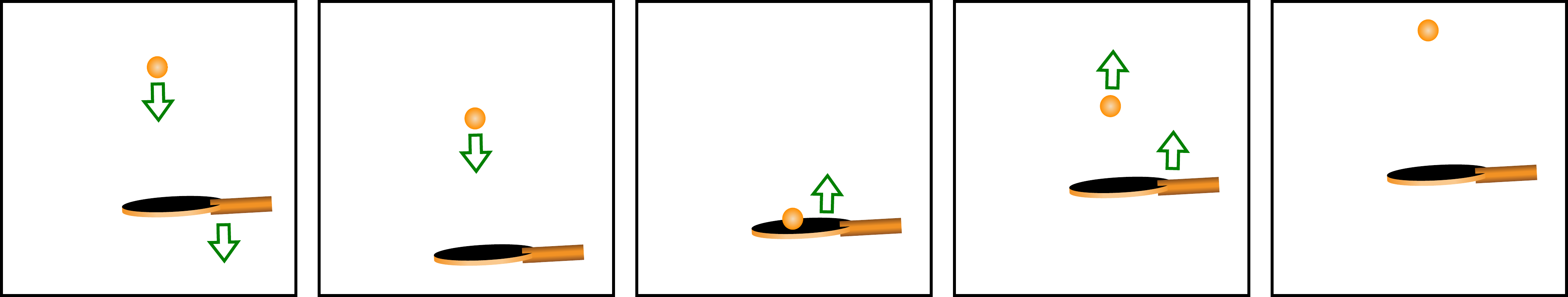}

}

\subfloat[\label{fig:simu_robo}Paddling movement for the simulated robot. The
black ball represents the imagined target (see Sect.~\ref{sub:Strategy})]{\includegraphics[width=1\textwidth]{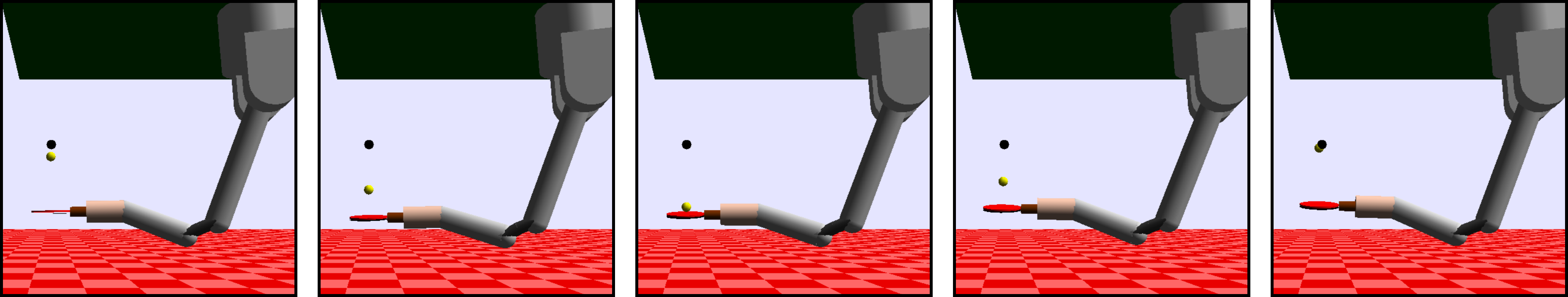}}

\subfloat[\label{fig:real_robot}Paddling movement for the real Barrett WAM.]{\includegraphics[width=1\textwidth]{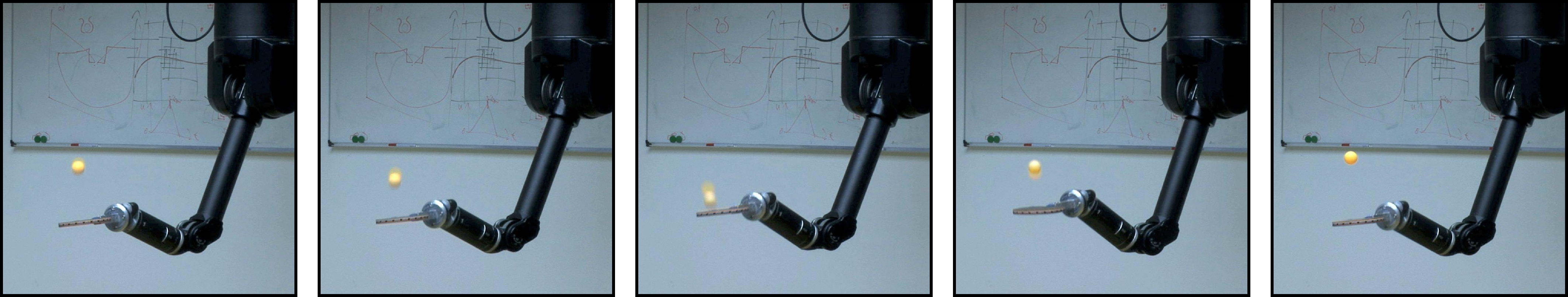}}\caption{\label{fig:bouncing-illu}This figure illustrates a possible sequence
of bouncing the ball on the racket in a schematic drawing, in simulation,
and on the real robot.}

\end{figure*}
The goal of the task is to bounce a table tennis ball above a racket.
The racket is held in the player's hand, or in our case attached to
the end-effector of the robot. The ball is supposed to be kept bouncing
on the racket. A possible movement is illustrated in Fig.~\ref{fig:bouncing-illu}.

It is desirable to stabilize the bouncing movement to a strictly vertical
bounce, hence, avoiding the need of the player to move a lot in space
and, thus, leaving the work space of the robot. The hitting height
is a trade-off between having more time until the next hit at the
expense of the next hitting position possibly being further away.
The task can be sub-dived into three intuitive primitives: hitting
the ball upward, moving the racket under the ball before hitting,
and changing the orientation of the racket to move the ball to a desired
location. A possible strategy is outlined in the next section.

The ball is tracked using a stereo vision setup and its positions
and velocities are estimated by a Kalman filter. To initialize the
ball-bouncing task, the ball is thrown towards the racket.

\subsection{Bouncing Strategy\label{sub:Strategy}}

\begin{figure}
\centering{}\includegraphics[width=0.7\columnwidth]{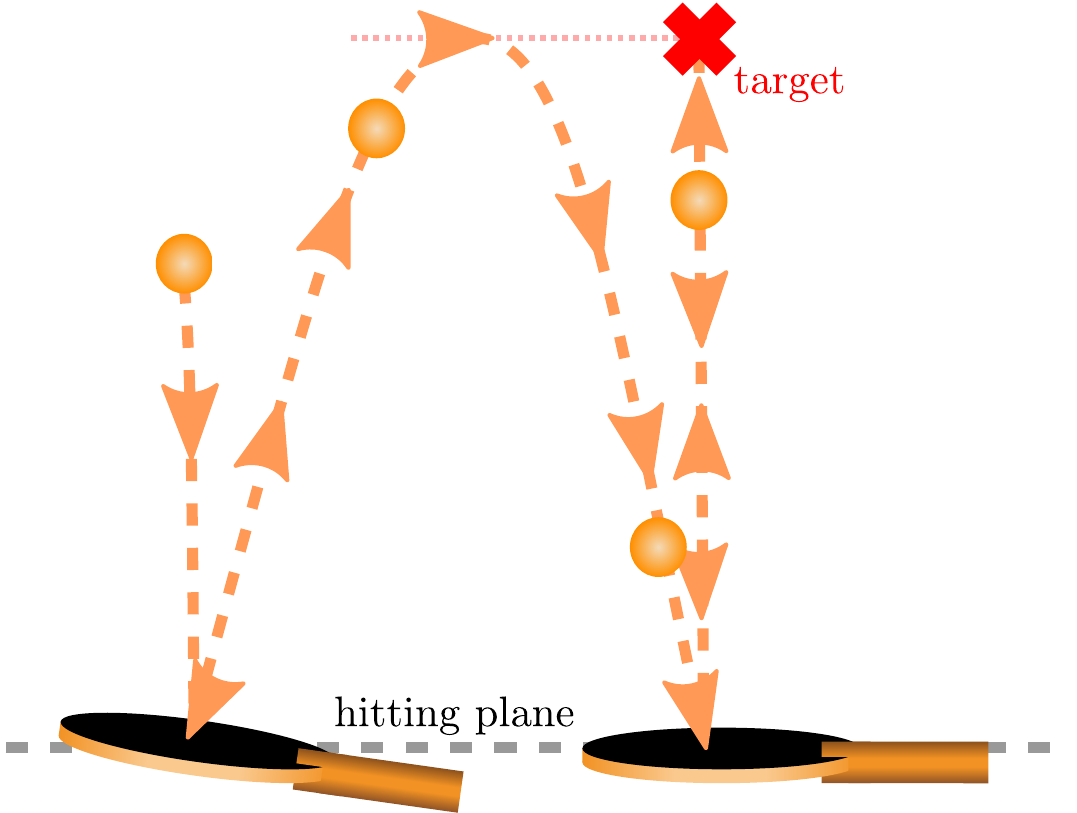}\caption{\label{fig:strategy}This figure illustrates the employed strategy
for bouncing the ball on the racket. The highest point of the ball
trajectory is supposed to coincide with the red target. The racket
is always hitting the ball in a fixed height, i.e., the hitting plane.
The strategy is to play the ball in a way that the next intersection
with the hitting plane is directly below the target and the maximum
height of the ball trajectory corresponds to the height of the target.
If the bounce works exactly as planned, the ball needs to be hit only
once to return to a strictly vertical bouncing behavior. }
\end{figure}

The strategy employed to achieve the desired bouncing behavior is
based on an imagined target that indicates the desired bouncing height.
This target is above the default posture of the racket. The top point
of the ball trajectory is supposed to hit this target, and the stable
behavior should be a strictly vertical bounce. This behavior can be
achieved by defining a hitting plane, i.e., a height at which the
ball is always hit (which corresponds to the default posture of the
racket). On this hitting plane, the ball is always hit in a manner
that the top point of its trajectory corresponds to the height of
the target and the next intersection of the ball trajectory with the
hitting plane is directly under the target. See Fig.~\ref{fig:strategy}
for an illustration. 

To achieve this desired ball behavior, the racket is always moved
to the intersection point of the ball trajectory and the hitting plane.
By choosing the hitting velocity and the orientation of the racket,
the velocity and direction of the ball after being hit can be changed.
The required hitting velocity and orientation are calculated using
a model of the ball and the racket. The ball is modeled as a point
mass that moves according to the ballistic flight equations. For the
relatively low speeds and small distances air resistance is negligible.
The contact with the racket is modeled as a reflection with a restitution
factor.

Using this strategy the ball can be brought back to a strictly vertical
bouncing behavior with a single hit. However, this method requires
the knowledge of the ball position and velocity, as well as a model
of the ball behavior. An alternative strategy that stabilizes the
behavior in a completely open loop behavior employs a slightly concave
paddle shape \cite{Reist2009}. A method similar to the proposed strategy
has been employed by\cite{Kulchenko2012,Muller2011}, and \cite{Buehler1994}
proposed the mirror law for this task. The ball bouncing task has
also be employed to study how humans stabilize a rhythmic task \cite{Schaal1996}.

\subsection{Learning Results\label{sub:Learning-Results}}

As discussed in Section~\ref{sub:Task-Description}, the task can
be described by three primitives: ``move under the ball'', ``hit
the ball'', and ``change racket orientation''. Training data is
collected in the relevant state space independently for each primitive.
For doing so, the parameters corresponding to the other primitives
are kept fixed and variants of the primitive are hence executed from
various different starting positions. The primitive ``move under
the ball'' corresponds to movements in the horizontal plane, the
primitive ``hit the ball'' to up and down movements, and the primitive
``change racket orientation'' only changes the orientation of the
end-effector. We collected 30 seconds of training data for each primitive,
corresponding to approximately 60 bounces.

Having only three primitives allows it to enumerate all six possible
dominance structures, to learn the corresponding prioritized control
law, and to evaluate the controller. As intuitive quality measure
we counted the number of bounces until the robot missed, either due
to imprecise control or due to the ball being outside of the safely
reachable work-space.

Table~\ref{tab:dominance} illustrates the resulting dominance structures.
The most relevant primitive is the ``hit the ball'' primitive, followed
by the ``move under the ball'' primitive. In the table it is clearly
visible that inverting the order of two neighboring primitives that
are in the preferred dominance order always results in a lower number
of hits. Compared to a single model, that was trained using the combined
training data of the three primitives, all but two prioritized control
laws work significantly better. The ordering may appear slightly counter-intuitive
as moving under the ball seems to be the most important primitive
in order to keep the ball in the air, allowing for later corrections.
However, the robot has a fixed base position and the ball moves quickly
out of the safely reachable work-space, resulting in a low number
of hits. Additionally, the default position of the racket is almost
vertical, hence covering a fairly large area of the horizontal plane
resulting in robustness with respect to errors in this primitive.

\begin{table}
\centering{}%
\begin{tabular}{|c|c|c|}
\hline 
\multirow{2}{*}{Dominance Structure} & \multicolumn{2}{c|}{Number of Hits}\tabularnewline
 & in Simulation & on Real Robot\tabularnewline
\hline 
\hline 
single model & $5.70\pm0.73$ & $1.10\pm0.99$\tabularnewline
\hline 
hit$\succeq$move$\succeq$orient & $11.35\pm2.16$ & $2.30\pm0.67$\tabularnewline
\hline 
hit$\succeq$orient$\succeq$move & $10.85\pm1.46$ & $1.70\pm0.95$\tabularnewline
\hline 
move$\succeq$hit$\succeq$orient & $9.05\pm0.76$ & $1.40\pm0.70$\tabularnewline
\hline 
move$\succeq$orient$\succeq$hit & $7.75\pm1.48$ & $1.40\pm0.84$\tabularnewline
\hline 
orient$\succeq$hit$\succeq$move & $5.90\pm0.85$ & $1.30\pm0.67$\tabularnewline
\hline 
orient$\succeq$move$\succeq$hit & $5.35\pm0.49$ & $1.30\pm0.48$\tabularnewline
\hline 
\end{tabular}\caption{\label{tab:dominance}This table shows the suitability of the possible
dominance structures (mean$\pm$std). The ``hit the ball'' primitive
clearly is the dominant one, followed by the ``move under the ball''
primitive. The prioritized control laws work significantly better
than a single model learned using the combined training data of the
three primitives. Preliminary results on the real robot confirm this
ordering.}
\end{table}

\addtolength{\textheight}{-10.5cm}   

\section{CONCLUSION\label{sec:Conclusion}}

In this paper, we have presented a prioritized control learning approach
that is based on the superposition of movement primitives. We have
introduced a novel framework for learning prioritized control. The
controls of the lower priority primitives are fulfilled as long as
they lay in the null-space of the higher priority ones and get overridden
otherwise. As representation for the primitives, we employ the dynamical
systems motor primitives \cite{ADM:Ijs2002,Schaal2007}, which yield
controls in the form of desired accelerations. These primitives are
executed separately to collect training data. Local linear models
are trained using a weighted regression technique incorporating the
various possible dominance structures. In the presented ball bouncing
task, the movement is restricted to a space where the controls are
approximately linear. Hence, a single linear model per primitive was
sufficient. This limitation can be overcome by either considering
multiple local linear models \cite{Peters2008a} or by kernelizing
the weighted regression, as described in Sect.~\ref{sub:Single-Control-Law}
and \ref{sub:Prioritized-Primitives-Control}.

The dominance structure of the task was determined by testing all
possible structures exhaustively. Intuitively, the lower priority
primitives represent a global behavior and the high priority primitives
represent specialized corrections, hence overriding the lower priority
controls. In most cases, the resulting prioritized control works significantly
better than a single layer one that was trained with the combined
training data of all primitives. As illustrated by the evaluations,
the dominance structure can have a significant influence on the global
success of the prioritized control. Enumerating all possible dominance
structures is factorial in the number of primitives and hence unfeasible
in practice for more than four primitives. In this case, smarter search
strategies are needed.

The success of the different dominance structures not only depends
on the task but also on the employed strategy of activating and adapting
the different primitives. An interesting area for future research
could be to jointly learn the prioritized control and the strategy.

The presented approach has been evaluated both in simulation and on
a real Barrett WAM and we have demonstrated that our novel approach
can successfully learn a ball-bouncing task.

\bibliographystyle{IEEEtran}
\bibliography{Kober2012LearningPrioritizedControl}

\end{document}